\documentclass[conference]{IEEEtran}
\IEEEoverridecommandlockouts
\usepackage{cite}
\usepackage{amsmath,amssymb,amsfonts}
\usepackage[ruled,vlined]{algorithm2e}
\usepackage{graphicx}
\usepackage{textcomp}
\usepackage{xcolor}
\usepackage[english]{babel}
\usepackage[utf8]{inputenc}
\usepackage{comment}
\usepackage{url}
\usepackage{lipsum}
\usepackage{hyperref}
\hypersetup{
    colorlinks=true,
    linkcolor=blue
}
\newcommand\blfootnote[1]{%
  \begingroup
  \renewcommand\thefootnote{}\footnote{#1}%
  \addtocounter{footnote}{-1}%
  \endgroup
}

\definecolor{lightblue}{HTML}{1F77B4}

\def\BibTeX{{\rm B\kern-.05em{\sc i\kern-.025em b}\kern-.08em
    T\kern-.1667em\lower.7ex\hbox{E}\kern-.125emX}}

\usepackage{subcaption}
\begin{document}

\title{Privacy Enhancement for Cloud-Based Few-Shot Learning}

\author{\IEEEauthorblockN{Archit Parnami\IEEEauthorrefmark{1},
Muhammad Usama\IEEEauthorrefmark{2},  
Liyue Fan\IEEEauthorrefmark{3} and
Minwoo Lee\IEEEauthorrefmark{4}}
\IEEEauthorblockA{Department of Computer Science\\
University of North Carolina at Charlotte, 
USA\\
Email: \IEEEauthorrefmark{1}aparnami@uncc.edu,
\IEEEauthorrefmark{2}msaleem2@uncc.edu,}
\IEEEauthorrefmark{3}liyue.fan@uncc.edu,
\IEEEauthorrefmark{4}minwoo.lee@uncc.edu}

\maketitle
\thispagestyle{plain}
\pagestyle{plain}

\begin{abstract}
Requiring less data for accurate models, few-shot learning has shown robustness and generality in many application domains.  However, deploying few-shot models in untrusted environments may inflict privacy concerns, e.g., attacks or adversaries that may breach the privacy of user-supplied data. This paper studies the privacy enhancement for the few-shot learning in an untrusted environment, e.g., the cloud, by establishing a novel privacy-preserved embedding space that preserves the privacy of data and maintains the accuracy of the model. We examine the impact of various image privacy methods such as blurring, pixelization, Gaussian noise, and differentially private pixelization (DP-Pix) on few-shot image classification and propose a method that learns privacy-preserved representation through the joint loss. 
The empirical results show how privacy-performance trade-off can be negotiated for privacy-enhanced few-shot learning.  
\end{abstract}


\blfootnote{Accepted in IEEE WCCI 2022 International Joint Conference on Neural Networks (IJCNN) @ Padua, Italy, July 2022.}

\section{Introduction}
There has been a widespread adoption of cloud-based machine learning platforms recently, such as Amazon Sagemaker \cite{joshi2020amazon}, Google AutoML \cite{bisong2019google}, and Microsoft Azure \cite{team2016azureml}. They allow  companies and application developers to easily build and deploy their AI applications as a Service (AIaaS).However, the users of AIaaS services may encounter two major challenges. 1)  \textbf{Large Data Requirement}: Deep Learning models usually require large amounts of training data. This training data  needs to be uploaded to the cloud services for the developers to build their models, which may  be inconvenient and infeasible at times.  2)  \textbf{Data Privacy Concerns}: Sharing data with untrusted servers may pose threats to end-user privacy. For instance, a biometric authentication
application deployed in the cloud will expose user photos to a third-party cloud service. 

\begin{figure}
    \centering
    \includegraphics[width=0.9\linewidth]{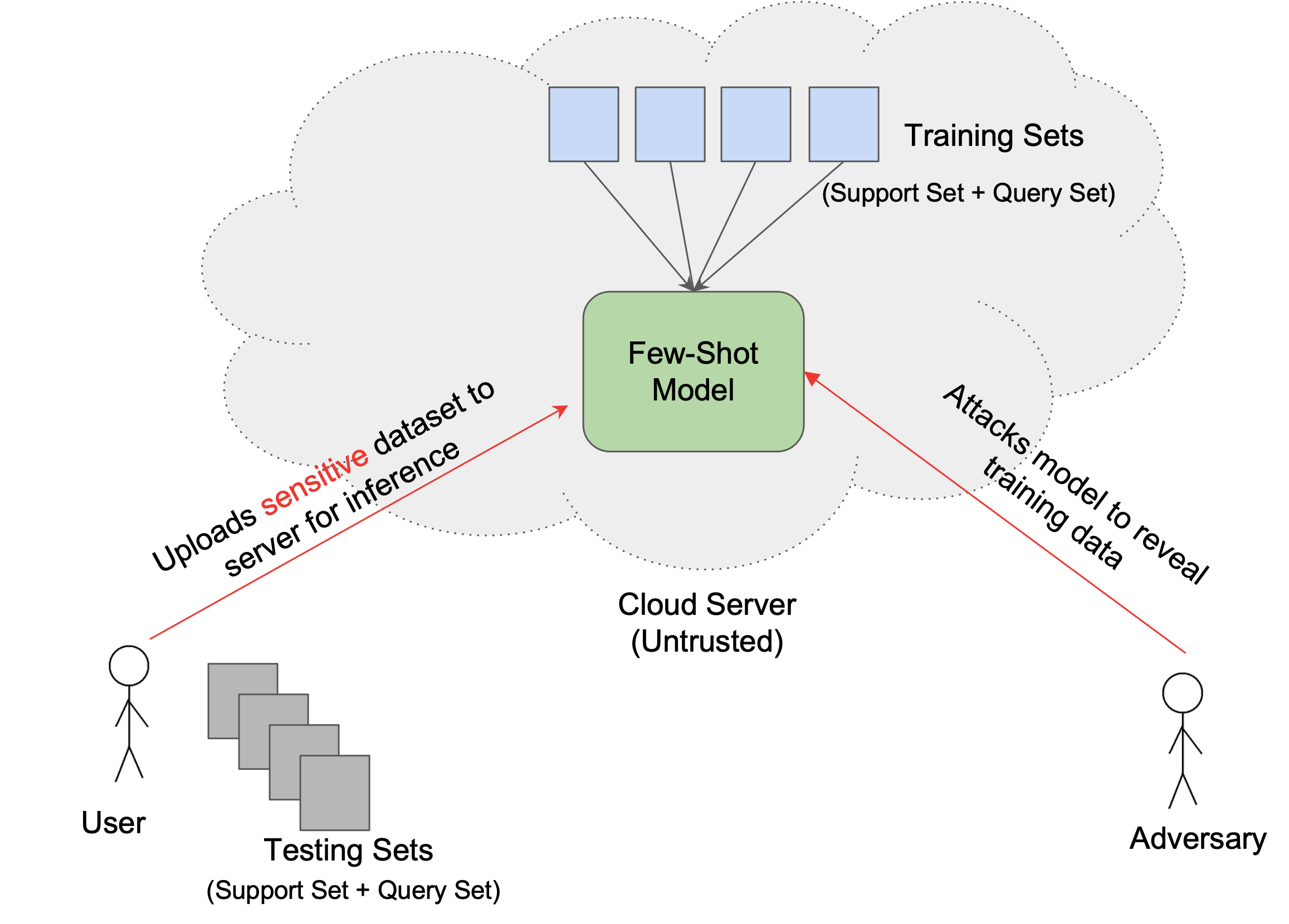}
    \caption{Threats in a cloud-based few-shot model. 1) attacks on training data~\cite{fredrikson2015model, shokri2017membership} and 2) exposure of sensitive dataset to untrusted cloud server for inference.}
    \label{fig:thread_model}
    \vspace{-0.75cm}
\end{figure}

To address the \textit{large data requirement} problem, there has been increasing research on the approaches that require less amount of training data, popularly known as Few-Shot Learning \cite{Wang2019FewshotLA}. Specifically, metric-based few-shot classification methods \cite{snell_prototypical_2017, vinyals_matching_2016, oreshkin_tadam:_2018, sung_learning_2017, yoon_tapnet:_2019} learn to map images of unseen classes into distinct embeddings that preserve the distance relation among them and then perform classification of the input query image by the distance to the class embeddings. Recent works have been able to achieve up to $\sim$90\% accuracy on the challenging task of 5-way 5-shot classification on the MiniImageNet dataset \cite{leaderboard}. Despite the success and promises of few-shot learning, it is imperative to address the \textit{data privacy concerns} to protect user-supplied sensitive data, e.g., when a metric-based few-shot model is deployed in a cloud server (Fig.~\ref{fig:thread_model}).

\begin{figure*}[ht!]
    \centering
    \includegraphics[width=0.9\textwidth]{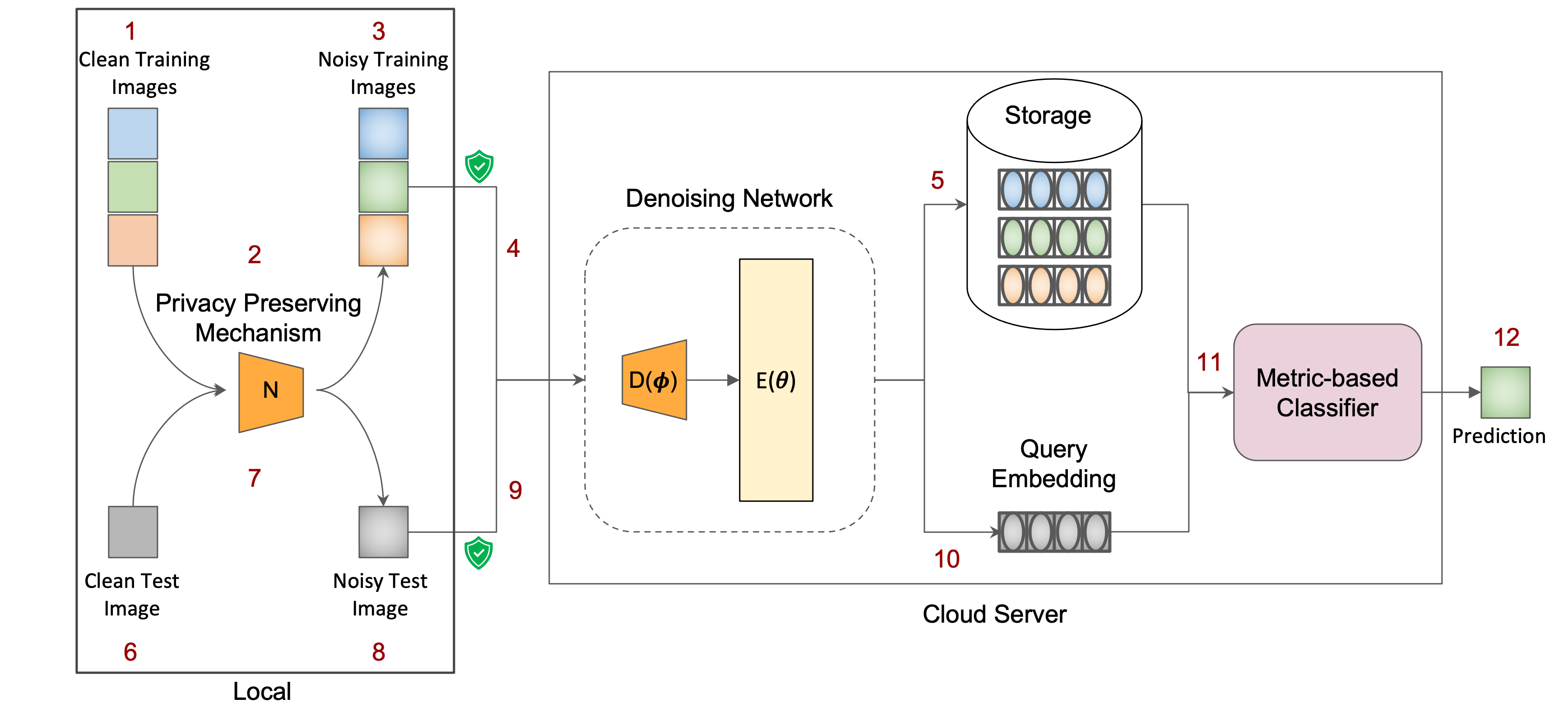}
    \caption{\textbf{Few-Shot Private Image Classification in the Cloud:} A denoising network is first trained with non-user data and deployed in the cloud.   Using a privacy preserving method (2), a user can obfuscate clean training images (1) to obtain noisy training images (3). These images are then sent to the cloud server where they are first denoised and then encoded (4) to be stored as privacy-preserved embeddings on the server (5).  A user can obfuscate the clean test image (6) and query the server using a noisy test image (8) to obtain a prediction (12).   
    }
    \label{fig: cloud}
    \vspace{-0.5cm}
\end{figure*}

Several privacy-preserving approaches may be adopted in machine learning applications, including cryptography, differential privacy, and data obfuscation.   Recent works~\cite{cabrero2021sok} adopted cryptographic techniques to protect the confidentiality of data. {For example, remote machine learning services can provide results on encrypted queries~\cite{cabrero2021sok}; a range of primitives, such as Homomorphic Encryption, may be adopted to manage the encrypted data.  Despite promising results, crypto-based methods inflict high computational overheads, creating challenges for practical deployment. Furthermore, such solutions may breach privacy by disclosing the exact computation results, and an adversary may utilize the model's output to launch inference attacks on training data~\cite{fredrikson2015model,shokri2017membership}. }
{Differential privacy~\cite{dwork2014privacybook} has been adopted to train machine learning models while providing indistinguishability guarantees for individual records in the training set~\cite{abadi_deep_2016}. However, the strong privacy guarantees tend to reduce the model performance and have shown disparate impacts on the underrepresented classes~\cite{disparateML}. In contrast, data obfuscation methods achieve privacy protection without inflicting high computational costs, e.g., image blurring and pixelization.  Obfuscation can be applied to protecting both training and testing data, and can provide differential privacy guarantees at individual-level data~\cite{fan_image_2018}. }

This paper focuses on the privacy of testing data (support+query) specifically for few-shot learning. A few-shot model built for clean images exhibits poor performance when tested with noisy/private image data. This is because meta-learning based few-shot models do not work well with out-of-distribution tasks \cite{finn_model-agnostic_2017, snell_prototypical_2017, parnami2022learning}. Therefore, applying the obfuscation methods to the image data and simply using an off-the-shelf pre-trained few-shot model leads to degradation in performance, as observed in our experiments (Fig. \ref{fig:results} Baseline Model). Hence, it is imperative to study privacy specifically in context of few-shot learning. To this end, we suggest a private few-shot learning approach trained on noisy data samples as illustrated in Fig.~\ref{fig: cloud}.  Adopting an obfuscation mechanism on the local input data samples, a user transfers privacy-encoded data to the cloud. The proposed jointly-trained, denoised embedding network, the \textit{Denoising Network}, constructs privacy-preserved latent space for robust few-shot classification. To validate the proposed approach,  we examine four privacy methods including traditional obfuscation methods such as Pixelization and Blurring, which do not provide quantifiable privacy guarantees~\cite{McPherson2016DefeatingIO}, and also Differentially Private Pixelization (DP-Pix) \cite{fan_image_2018} which provides differential privacy guarantees.  

This study examines practical implications for a holistic private few-shot learning framework on an untrusted service platform, which has not been studied previously.  Thus, our main contributions are 1) first proposing a unified framework for deploying few-shot learning models in the cloud while protecting the privacy of user-supplied sensitive data and 2) thoroughly examining privacy methods on three different datasets of varying difficulty, therefore 3) discovering and observing the existence of the effective privacy-preserved latent space for few-shot learning.

\section{Few-Shot Learning}
Few-shot learning is a subfield of machine learning that focuses on the ability of machine learning models to generalize from few-training examples. The recent progress in the field has largely come from a machine learning technique called meta-learning \cite{finn_model-agnostic_2017}. The idea is to train a model on numerous different but similar kinds of tasks such that the model can generalize on a new test task (as long as the test task is from the same distribution of the tasks the model was initially trained on). There are three kinds of Few-shot learning: Metric-based, optimization-based, and model-based \cite{vinyals}. Here we only discuss metric-based techniques and refer the reader to our survey paper \cite{parnami2022learning} for further reading. 

The early nominal works in metric-based few-shot learning methods are: Prototypical Networks \cite{snell_prototypical_2017}, Matching Networks \cite{vinyals_matching_2016}, Relation Networks \cite{sung_learning_2017} etc.  In all these methods, the network learns to encode embeddings of input images such that images that belong to same class are closer to each other and those from different class are farther apart, where the  idea of closeness is defined in terms of a metric such as euclidean.

\subsection{Few-Shot Classification} \label{FSC}
We base our framework (Fig.~\ref{fig: cloud}) on Prototypical Networks \cite{snell_prototypical_2017} for building our \textbf{Few-Shot Private Image Classification (FS-PIC)} model. The model is trained on a labeled dataset $D_{train}$ and tested on $D_{test}$. The set of classes present in $D_{train}$ and $D_{test}$ are disjoint. The test set has only a few labeled samples per class. We follow an episodic training paradigm in which each episode the model is trained to solve an $N$-way $K$-Shot private image classification  task. Each episode $e$ is created by first sampling $N$ classes from the training set and then sampling two sets of examples from these classes: (1) the support set $S_{e} = \{(s_{i},y_{i})\}_{i=1}^{N \times K}$ containing $K$ examples for each of the $N$ classes and (2) the 
query set $Q_{e} = \{(q_{j} , y_{j} )\}_{j =1}^{N \times H}$ containing $H$ different examples from the same $N$ classes. The episodic training for the FS-PIC task  minimizes, for each episode, the loss on the prediction of samples in the query set, given the support set. The model is a parameterized function, and the loss is the negative log-likelihood of the true class of each query sample:

\begin{equation}\label{eq:loss}
\mathcal{L}(\theta) =    -\sum_{t=1}^{\vert Q_{e} \vert} \log P_{\theta} (y_{t} \mid q_{t}, S_{e}),
\end{equation}
where $(q_{t}, y_{t}) \in Q_{e}$ is the sampled query, $S_{e}$ is the support set at episode $e$, and $\theta$ is the model parameter.

Prototypical Networks make use of the support set to compute a centroid (prototype) for each class (in the sampled episode) and query samples are classified based on the distance to each prototype. For instance, a CNN  $f : \mathbb{R}^{n_{v}} \to \mathbb{R}^{n_{p}}$, parameterized by $\theta_{f}$, learns a $n_{p}$-dimensional space where $n_v$-dimensional input samples of the same class are close and those of different classes are far apart. For every episode $e$, each embedding prototype $p_{c}$ (of class $c$) is computed by averaging the embeddings of all support samples of class $c$ as
\begin{equation} 
    p_{c} = \frac{1}{K} \sum_{(s_{i}, y_{i}) \in S_{e}^c} f(s_{i}), 
\end{equation}
where $S_{e}^c \subset S_{e}$ is the subset of support examples belonging to class c. Given a  distance function $d$, the distance of the query $q_{t}$ to each of the class prototypes  $p_{c}$ is calculated. By taking a softmax \cite{bridle1990probabilistic} of the measured (negative) distances, the model produces a distribution over the $N$ classes in each episode:
 \begin{equation}
    P(y = c \mid q_{t}, S_{e}, \theta)  = \frac{exp(-d(f(q_{t}),p_{c}))}{\sum_{n} exp(-d(f(q_{t}),p_{n}))},
\end{equation}
where  metric $d$ is a Euclidean distance and the parameters $\theta$ of the model are updated with stochastic gradient descent by minimizing Equation~(\ref{eq:loss}). Once the training finishes, the parameters $\theta$ of the network are frozen. Then, given any new FS-PIC task, the class corresponding to the maximum $P$ is the predicted class for the input query $q_{t}$.

\section{Privacy Methods} \label{privacy_methods}
We study following methods to introduce privacy in images (depicted in Fig. \ref{fig:privacy_methods} of Appendix).

\subsection{Independent Gaussian Noise}
Introducing some noise in an image is one way to distort information \cite{Mivule2013UtilizingNA}. Kim \cite{Kim2002AMF}, first publicized the work on additive noise by the general expression $ Z = X + \epsilon $, where $X$ is the original data point, $\epsilon$ is the random variable (noise) with a distribution $\epsilon \sim  \mathcal{N}(0,\sigma^2)$ and $Z$ is the transformed data point, obtained by the addition of noise $\epsilon$ to the input $X$.

Therefore, for an image with dimensions $(H, W, C)$, we sample $H \times W \times C$ values from a Gaussian (normal) distribution with mean ($\mu$) zero and standard deviation $\sigma$ of the probability density function
$p(x) = \frac{1}{\sqrt{2\pi\sigma^2}}\exp{-\frac{(x-\mu)^2}{2\sigma^2}}$.
We use the implementation from \cite{harris2020array}.

\subsection{Common Image Obfuscation}
Two widely used image obfuscation techniques are \textit{Pixelization} and \textit{Blurring}.

\paragraph{Pixelization~\cite{Hill2016OnT}}  (also referred to as mosaicing) can be achieved by superposing a rectangular grid of size $b \times b$ over the original image and averaging the color values of the pixels within each grid cell. 
    
\paragraph{Blurring} i.e., Gaussian blur, removes details from an image by convolving a 2D Gaussian kernel with the image. Let the radius of blur be  $r$, then the size of the 2D kernel is given by $ (2r+1) \times (2r+1)$. Then, the values in this 2D kernel are sampled from the distribution: 
    
    \begin{equation}
        G(x,y) = \frac{1}{\sqrt{2\pi\sigma^2}}\exp{-\frac{(x^2 + y^2)}{2\sigma^2}},
    \end{equation}
    
    where $(x,y)$ are the coordinates inside the 2D kernel with origin at the center and the standard deviation $\sigma$ is approximated from the radius $r$ \cite{Gwosdek2011TheoreticalFO}. We use Pillow Image Library \cite{clark2015pillow} for the implementation.

\subsection{Differentially Private Image Pixelization}
Differential privacy (DP) is the state-of-the-art privacy paradigm for statistical databases~\cite{dwork2014privacybook}. Differentially Private Pixelization (DP-Pix) \cite{fan_image_2018} extends the DP notion to image data publication. It introduces a concept of $m$-Neighborhood, where two images ($I_1$ and $I_2$) are neighboring images if they differ by at most $m$ pixels. By differential privacy, content represented by up to $m$ pixels can be protected. A popular mechanism to achieve DP is the Laplace mechanism. However, the global sensitivity of direct image perturbation would be very high i.e., $\Delta I = 255m$, leading to high perturbation error.  The DP-Pix method first performs pixelization $P_b$ (with grid cells of $b \times b $ pixels) on the input image $I$, and then applies Laplace perturbation to the pixelized image $P_b(I)$, effectively reducing the sensitivity $\frac{255m}{b^2}$. The following equation summarizes the algorithm ($\tilde{P_b}$) to achieve $\epsilon$-differential privacy:
\begin{equation}
        \tilde{P_b}(I) = P_b(I) + {L}_{p},
\end{equation}
where each value in $L_{p}$ is randomly drawn from a Laplace distribution with mean 0 and scale $\frac{255m}{b^2\epsilon}$.  The parameter $\epsilon >0$ specifies the level of DP guarantee, where smaller values indicate stronger privacy. As DP is resistant to post-processing~\cite{dwork2014privacybook}, any computation performed on the output of DP-Pix, i.e., the perturbed pixelized images, would not affect the $\epsilon$-DP guarantees. Our approach proposes a denoising module for the obfuscated images by DP-Pix, improving the latent representation without sacrificing DP guarantees.

\section{Privacy Enhanced Few-shot Image Classification}
To build a few-shot model that can preserve the privacy of the input images, we can utilize any of the privacy methods discussed in the previous section. However, doing so may degrade the few-shot classification performance tremendously. To avoid this, we introduce a denoiser and train it jointly for few-shot classification using meta-learning on noisy images (Fig. \ref{fig: cloud}). Together, the denoiser and the embedding network forms our \textit{Denoising Network}. Combined with the properly chosen privacy method, the Denoising Network aims to discover a privacy-preserved latent embedding space (not denosing to recover the original image), where \textbf{the privacy of input data is be preserved} and \textbf{robustness and generality for few-shot classification are maintained}.

\noindent \textbf{Denoiser: } Zhang et al. \cite{zhang_beyond_2017} proposed a denoising convolutional neural network (DnCNN) which uses residual learning to output Gaussian noise.  Specifically, the input of the network is a noisy observation such that $y = x + v$ where $y$ is the input image, $x$ be the clean image, and $v$ be the actual noise. The network learns the residual mapping function $\mathcal{R}(y) \approx v$ and predicts the clean image using $x = y - \mathcal{R}(y)$. The averaged mean squared error between the predicted residue and actual noise is used as the loss function to train this denoiser with parameters $\phi$ as
\begin{equation}
    \mathcal{L}(\phi) = \frac{1}{2N}\sum_{i=1}^N || \mathcal{R}(y_i; \phi) - (y_i - x_i) ||^2.
\end{equation}
We plug the DnCNN denoiser into our FS-PIC pipeline (Fig. \ref{fig: cloud}) to estimate the clean image before pixelization, blurring, Gaussian noise, and DP-Pix. The architecture for the denoiser can be found in Fig. \ref{fig:denoiser} of Appendix.

\noindent \textbf{Embedding Network: } Partially denoised images from the denoiser $D(\phi)$ are fed to embedding network $E(\theta)$ to obtain denoised embeddings,  which then form the class prototypes. The classification loss is measured using Eq. ~\ref{eq:loss}. 

\noindent The total loss for training the \textit{Denoising Network} (Denoiser + Embedding Network) is formulated as the sum of denoising loss and classification loss:

\begin{equation} \label{eq:total_loss}
    \mathcal{L} =  \mathcal{L}(\phi) +  \mathcal{L}(\theta).
\end{equation}
The joint loss enforces the reduction of noise in input images while learning the distinctive representations that maximize the few-shot classification accuracy. This simple loss guides the embedding space towards privacy-preserved latent space without losing its generality. For Prototypical Networks, the prototypes are expected to be the centers of the privacy-preserved embeddings for each class.  Although the sum of losses can be weighted, our experiments observed that weighting did not significantly impact the final accuracy of the few-shot image classification model as long as the weighting coefficients are non-zero. We outline the  episodic training process used for building a FS-PIC model in Algorithm \ref{alg:episodic} and describe the notations used in Table \ref{symbols}. 

\begin{table}
\centering
\begin{tabular}{p{0.2\linewidth} | p{0.7\linewidth}}
\hline
Notation  & Description  \\
\hline
$t$ & \#examples in the training set \\
$M$ & \#classes in the training set \\
$N <= M$  & \#classes sampled per episode \\
$K$  & \#support examples sampled per class \\
$H$ & \#query examples sampled per class  \\
\hline
\end{tabular}
\caption{Symbols}
\label{symbols}
\end{table}

\begin{algorithm}

\DontPrintSemicolon
\SetAlgoLined


\textbf{Input:} $D = \{(x_1, y_1), ...,  (x_t, y_t)\}$  where $y_i \in \{1,...,M\}$. $D^c$ denotes the subset of D containing all elements $(x_i, y_i)$ such that $y_i = c$.
\BlankLine

\While {True}
{   
    \tcp{Select a set of N classes}
    $V \leftarrow \textrm{RandomSample}(\{1,...,M\}, N)$ \\
  
    \For{$c$ in $V$}  
    {
       
    \tcp{Select support examples}
    $S^{c}_{e} \leftarrow \textrm{RandomSample}(D^c, K)$ \\
    
    \tcp{Select query examples}
    $Q^{c}_{e} \leftarrow \textrm{RandomSample}(D^c \setminus S^{c}_{e}, H)$ \\
    
    \tcp{Add noise} 
    $\hat{S^{c}_{e}} \leftarrow \textrm{AddNoise}(S^{c}_{e}, \epsilon)$ \\
    $\hat{Q^{c}_{e}} \leftarrow \textrm{AddNoise}(Q^{c}_{e}, \epsilon)$ \\
    }

    \tcp{Form a set of all  clean images}
    $S_e \leftarrow \{S^{1}_{e}, S^{2}_{e}, ... S^{N}_{e}\}$ \\
    $Q_e \leftarrow \{Q^{1}_{e}, Q^{2}_{e}, ... Q^{N}_{e}\}$ \\
    $X_e \leftarrow \{S_e, Q_e\}$
    
    \tcp{Form a set of all noisy images}
    $\hat{S_e} \leftarrow \{\hat{S^1_{e}}, \hat{S^2_{e}}, ... \hat{S^N_{e}}\}$ \\
    $\hat{Q_e} \leftarrow \{\hat{Q^1_{e}}, \hat{Q^2_{e}}, ... \hat{Q^N_{e}}\}$ \\
    $\hat{X_e} \leftarrow \{\hat{S_e}, \hat{Q_e}\}$ \\
    
    \tcp{Apply the denoiser}
    $\bar{X_e} \leftarrow G(\hat{X_e}; \theta)$ \\
    $\bar{S_e}, \bar{Q_e} \leftarrow \bar{X_e}$ \\  
  
    \tcp{Calculate denoising loss}
    $\mathcal{L}_d \leftarrow \textrm{MSE}(\bar{X_e}, X_e)$ \\

   \tcp{Compute class prototypes using denoised support examples}
   \For{$c$ in $V$}  
    {
      $\bar{p_c} \leftarrow \dfrac{1}{K} \sum\limits_{(\bar{x_i}, y_i) \in \bar{S^{c}_{e}}} f_{\phi}(\bar{x_i})$
    }
    
    $\mathcal{L}_c \leftarrow 0$ \\

    \For{$c$ in $V$}
    {
        \For{$(\bar{x_i},y_i)$  in $\bar{Q^{c}_{e}}$}
        {
            $\mathcal{L}_c  \leftarrow \mathcal{L}_c  + \frac{1}{N H} [ d(f_{\phi}(\bar{x_i}), \bar{p_c}) + \log \sum\limits_{c'} \exp(-d(f_{\phi}(\bar{x_i}), \bar{p_c})) ]$ \\
        }
    }

    $\mathcal{L} \leftarrow  \mathcal{L}_d +  \mathcal{L}_c$ \\

    $\phi \leftarrow \phi - \alpha_\phi \frac{\partial \mathcal{L}}{\partial \phi}$ \\
    $\theta \leftarrow \theta - \alpha_\theta \frac{\partial \mathcal{L}}{\partial \theta}$ \\
}

\caption{FS-PIC model training}
\label{alg:episodic}

\end{algorithm}

\section{Experiments}
\noindent \textbf{Datasets:} \textbf{1)} \textit{Omniglot} \cite{lake_one_2011} is a dataset of 1623 handwritten characters collected from 50 alphabets. Each character has 20 examples drawn by a different human subject.  We follow the same procedure as in \cite{vinyals_matching_2016} by resizing the gray-scale images to $28 \times 28 $ and augmenting the character classes with rotations in multiples of 90 degrees. Our training, validation, and testing split is of sizes 1028, 172, and 423 characters, respectively (or $4\times$ with augmentation). \textbf{2)} \textit{CelebFaces Attributes Dataset (CelebA)} \cite{liu2015faceattributes} is a large-scale face attributes dataset with more than 10K celebrity (classes) images. For the purpose of our experiments, we select classes that have at least 30 samples. This gives us 2360 classes in total, out of which 1510 are used for training, 378 for validation, and 427 for testing. We use aligned and cropped version of the dataset in which images are of dimension $218 (h) \times 178 (w)$. We center crop each image to $176 \times 176$ and then resize to $84 \times 84$. \textbf{3)} \textit{MiniImageNet} \cite{vinyals_matching_2016} dataset contains 100 general object classes where each class has 600 color images. The images are resized to $84 \times 84$, and the dataset is split into 64 training, 16 validation, and 20 testing classes following \cite{snell_prototypical_2017}.

\noindent \textbf{Settings for Privacy Methods:}
We explore the following parameters for each privacy method.  
Gaussian Blur with radius  $r = \{1,2,3,4,5\}$ is used for blurring images.
A filter window of size $b \times b$ where $b = \{2,4,6,8,10\}$ is used for pixelization. The pixelated image is then resized to match the model input dimensions. 
We perform experiments with Gaussian noise $\epsilon \sim  \mathcal{N}(\mu, \sigma)$ with mean $\mu$ = 0 and standard deviation $\sigma = \{40,80,120,160,200\}$.
For DP-Pix, we fix $\epsilon = 3$, $m = 1$ and vary pixelization parameter $b$ with values $\{2,4,6,8,10\}$. 

\noindent \textbf{Denoising Network:}
We use a lighter version of the DnCNN \cite{zhang_beyond_2017} model i.e., with 8 CNN layers instead of 17, for first denoising the image and subsequently feeding the denoised image into one of the following embedding networks.
\textbf{Conv-4} is a 4-layered convolutional neural network with 64 filters in each layer originally proposed in \cite{snell_prototypical_2017} for few-shot classification.
\textbf{ResNet-12} is a 12-layer CNN with 4 residual blocks. It has been shown to have better classification accuracy on few-shot image classification tasks. The architecture of the two embedding networks are detailed in Fig. \ref{fig:encoders} of Appendix.

\noindent \textbf{Training and Evaluation:}
We train using N-way K-shot PIC tasks (Algorithm. \ref{alg:episodic}) and use Adam optimizer
 with learning rate $\alpha_\theta = \alpha_\phi = 0.001$ with a decay of 0.5 every 20 epochs. Table \ref{table:training_parameters} lists the hyperparameters for the three datasets. The network is trained to minimize total loss of denoiser and classifier (Eq. \ref{eq:total_loss}). 
  We evaluate the performance by sampling 5-way 5-shot PIC tasks (with same privacy settings) from the test sets and measure the classification accuracy. The final results report the performance averaged over 1000 test episodes for the Omniglot dataset, and 600 test episodes for both MiniImageNet and CelebA datasets.
  To measure the effectiveness of the proposed denoising embedding space, we both train and evaluate each model's performance in two settings: 1) \textbf{without using the denoiser} and 2) \textbf{jointly training the denoiser with the classifier} i.e., the proposed \textit{ Denoising Network}.

\begin{table}[]
\centering
\begin{tabular}{|l|c|c|c|}
\hline
         & Omniglot & CelebA  & MiniImageNet \\ \hline
Way      & 60       & 5       & 5       \\ \hline
Shots    & 5        & 5       & 5       \\ \hline
Query    & 5        & 5       & 15      \\ \hline
Epochs   & 500      & 200     & 200     \\ \hline
Patience & 50       & 20      & 20      \\ \hline
Episodes & 100      & 100     & 100     \\ \hline
\end{tabular}
\caption{Training Hyperparameters}
\label{table:training_parameters}
\end{table}

\noindent \textbf{Privacy Risk Evaluation:}\label{sec:priv_desc} Privacy attacks on trained models such as model inversion\cite{fredrikson2015model} and membership inference\cite{shokri2017membership} are not applicable in our setting because the denoising and embedding models are trained with publicly available classes (data) using meta-learning.  The user-supplied test data (support and query set) are obfuscated for privacy protection. 
A practical privacy attack on obfuscated images is to infer the identities using existing facial recognition systems and public APIs, e.g., Rekognition. In this study, our goal is to investigate (1) the efficacy of the studied image obfuscation methods for privacy protection and (2) whether the proposed denoising approach has effects on privacy.  To simulate a powerful adversary, we apply the state-of-the-art face recognition techniques, e.g., FaceNet with the Inception ResNet V1 network\cite{facenet}, on the CelebA dataset; MTCNN \cite{MTCNN} is applied to detect and resize the facial region in each input image.  Specifically, 1000 entities were randomly selected from the CelebA dataset. For each entity, we randomly sampled 30 images, which were then partitioned between training and testing (20 : 10).  Different versions of the test set were generated by applying image obfuscation methods with various parameter values (denoted as \texttt{Noisy}) and by applying the proposed Denoising Network (denoted as \texttt{Denoised}).  We fine-tuned the Inception network and trained an SVC classifier on the clean training data.  In Fig. \ref{fig:reidentificationresults}, we report the accuracy on the noisy and denoised test sets, i.e., success of re-identification, with higher values indicating higher privacy risks.

\begin{figure*}[ht]
    \centering
    \includegraphics[width=0.95\textwidth]{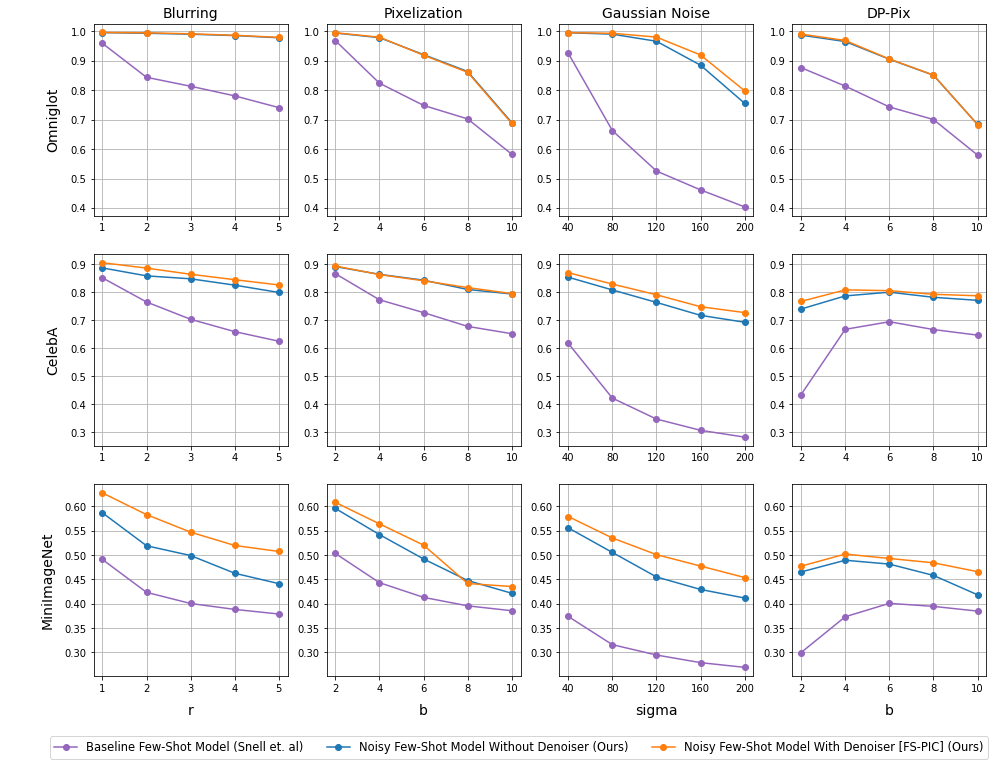}
    \caption{Test accuracy (y-axis) of 5-way 5-shot private image classification tasks sampled from Omniglot (top), CelebA (center) and MiniImageNet (bottom) datasets, presented with different privacy settings (x-axis) when using Conv-4 as encoder.}
    \label{fig:results}
\end{figure*}

\begin{figure*}[ht]
    \centering
    \includegraphics[width=0.95\textwidth]{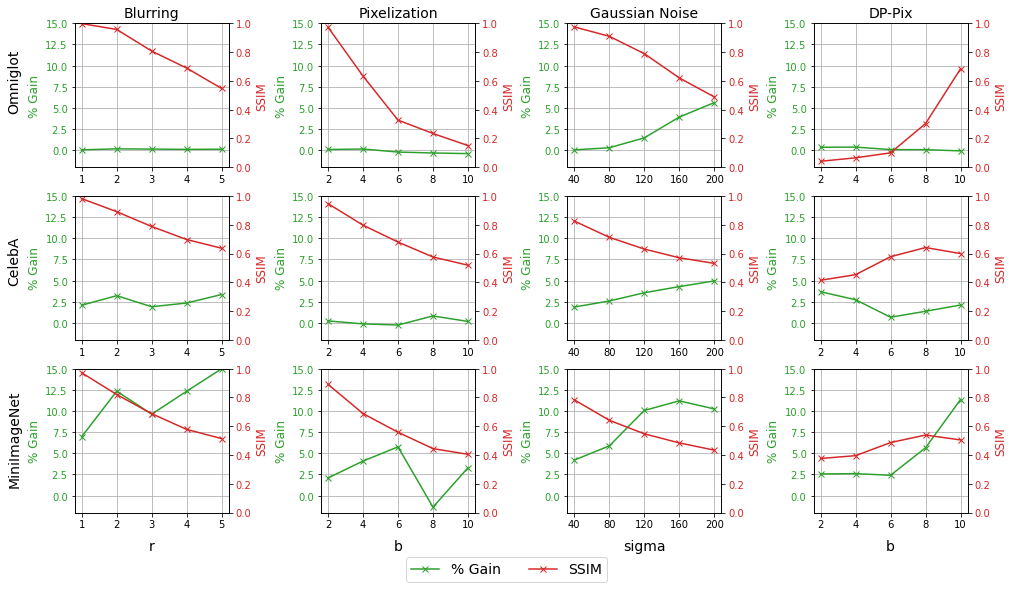}
    \caption{\% Gain vs SSIM for Conv-4  }
    \label{fig:gain_vs_ssim}
\end{figure*}

\section{Results and Discussions}
\begin{table}[h]
\centering
\begin{tabular}{|c|c|c|c|}
\hline
          & Omniglot & CelebA & MiniImageNet \\ \hline
Conv-4    & 0.99     & 0.90         & 0.61   \\ \hline
ResNet-12 &   --       & 0.92         & 0.65   \\ \hline
\end{tabular}
\caption{Baseline test accuracy of 5-way 5-shot classification of clean images. Omniglot is not evaluated for ResNet-12 because of its already near 100\% performance. }
\label{Tab:Baseline}
\end{table}

\subsection{Task Difficulty} The average 5-way 5-shot classification accuracy of our baseline few-shot model \cite{snell_prototypical_2017} trained on clean images and tested on clean images is 99\% on Omniglot dataset, 91\% on CelebA dataset, and 61\% on MiniImageNet dataset using Conv-4 encoder (Table \ref{Tab:Baseline}). This shows the approximate level of difficulty of few-shot tasks for each dataset i.e., Omniglot tasks are easy, tasks from CelebA have medium difficulty, and MiniImageNet tasks are hard. 

\subsection{Generalization}
We compare results for few-shot private image classification using three models in Fig.~\ref{fig:results}: 
\begin{enumerate}
    \item \textbf{Baseline Few-Shot Model:} When the few-shot model is trained on clean images and is tested on noisy images. 
    \item \textbf{Noisy Few-Shot Model Without Denoiser: } When the baseline few-shot model is trained on noisy images and is tested on noisy images with same privacy settings.
    \item \textbf{Noisy Few-Shot Model With Denoiser: } When the baseline few-shot model is \textit{jointly trained with the denoiser} on noisy images and is tested on noisy images with same privacy settings (Algorithm \ref{alg:episodic}).
\end{enumerate}

In all cases, we observe that noisy few-shot models outperforms the baseline few-shot model with wide gap. Also, in most cases, we note that adding a denoiser improves the accuracy. To better observe the effectiveness of denoiser, in Fig.~\ref{fig:gain_vs_ssim}, we quantify the improvement by calculating $\text{\% Gain} = \frac{\text{accuracy with denoiser} - \text{accuracy without denoiser}}{\text{accuracy without denoiser}} \times 100$.  We also quantify the change to the original image caused by the privacy method (post denoising) by calculating Structural Similarity Index (SSIM) \cite{SSIM} between denoised image and original clean image, averaged over 100 test images for each dataset and privacy parameter. 

\textbf{Blurring, Pixelization and Gaussian Noise:} As we increase the value of privacy parameters, the SSIM decreases, suggesting the higher dissimilarity between the denoised images and the original image (Fig \ref{fig:gain_vs_ssim}). Despite the degradation caused by the privacy method to the original image, we observe positive \% Gains for all three datasets. Specifically, on hard tasks (MiniImageNet), a gain of upto 15\% in accuracy ($r=5$) with the proposed \textit{Denoising Network}, reaffirming the generality of the few-shot learning. For easy (Omniglot) and medium (CelebA) tasks, where the baseline accuracy is already high, a relatively small positive gain of up to 5\% ($\sigma=200$) is reported.

\textbf{DP-Pix:} As we increase the size of pixelization window ($b$), the amount of Laplace noise that we add to the image decreases (as defined by $\frac{255m}{b^2\epsilon}$); however, the image quality decreases because of increasing pixelization. Therefore, we observe a trade-off point where the accuracy first increases and then decreases as we increase $b$ (Fig \ref{fig:results}). This trade-off is particularly observed for CelebA and MiniImageNet datasets. For Omniglot dataset, the performance just decreases with increasing $b$ because of the low resolution images in the dataset. From Fig. \ref{fig:gain_vs_ssim}, we observe that DP-Pix has the lowest SSIM values when compared with other privacy methods causing the most notable changes on the original image. 
Interestingly, we note that even with low SSIM values, we find instances that exhibit moderate \% gain i.e., at $b={2,4,6}$ indicating the presence of privacy preserving denoising embeddings. Results using ResNet-12 encoder are presented in Fig. \ref{fig:results_resnet} and \ref{fig:gain_resnet} of Appendix.

\subsection{Empirical Privacy Risks}
\begin{figure}[h!]
       \centering
        \begin{subfigure}[b]{0.95\columnwidth}  
        \centering
        \includegraphics[width=\textwidth]{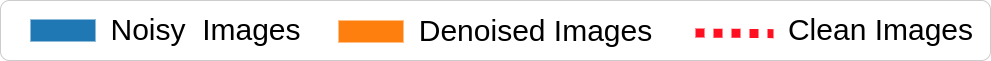}
        \caption*{}
        \vspace{-0.5cm}
    \end{subfigure}

        \begin{subfigure}[b]{0.45\columnwidth}  
        \centering
        \includegraphics[width=\textwidth]{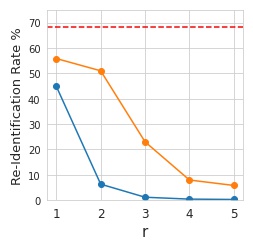}
        \caption[]%
        {{\scriptsize Blurring}}\label{fig:reid:blur}
    \end{subfigure}
    \begin{subfigure}[b]{0.45\columnwidth}
        \centering
        \includegraphics[width=\textwidth]{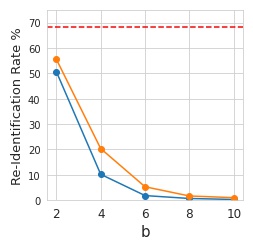}
        \caption[]%
        {{\scriptsize Pixelization}}\label{fig:reid:pix}
    \end{subfigure}

        \begin{subfigure}[b]{0.45\columnwidth}  
        \centering
        \includegraphics[width=\textwidth]{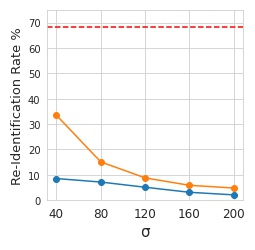}
        \caption[]%
        {{\scriptsize Additive Gaussian}}\label{fig:reid:gauss}
    \end{subfigure}
    \begin{subfigure}[b]{0.45\columnwidth}
        \centering
        \includegraphics[width=\textwidth]{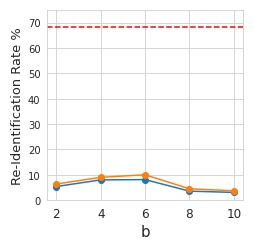}
        \caption[]%
        {{\scriptsize DP-Pix ($\epsilon=3$)}}\label{fig:reid:dp_pix}
    \end{subfigure}
    \caption{\small Privacy Risk Evaluation with CelebA}
    \label{fig:reidentificationresults}
    \vspace{-0.5cm}
\end{figure}

As described earlier, this experiment showcases the efficacy of image obfuscation methods against a practical adversary who utilizes state-of-the-art face recognition models trained on clean images. Furthermore, this experiment investigates whether the proposed denoising network has effects on privacy protection empirically. To this end, we vary the algorithmic parameters of the privacy methods and report the face re-identification rates on both obfuscated and denoised images in Fig.~\ref{fig:reidentificationresults}. For the clean test set sampled from CelebA, the re-identification accuracy is 68.12$\%$. 

\textbf{Blurring:}
In Fig.~\ref{fig:reid:blur}, the privacy risks are quite high with Gaussian kernel size $r=1$ for both blurred and denoised images.  As we increase the radius $r$, the chance of re-identifying the image decreases rapidly.  We also observe that after denoising, the blurred images are more likely to be re-identified. For instance, at $r=2$, the re-identification rate is 7.34\% for blurred images but 54.01\% for denoised images.

\begin{figure*}[h!]
    \centering
    \includegraphics[width=\textwidth]{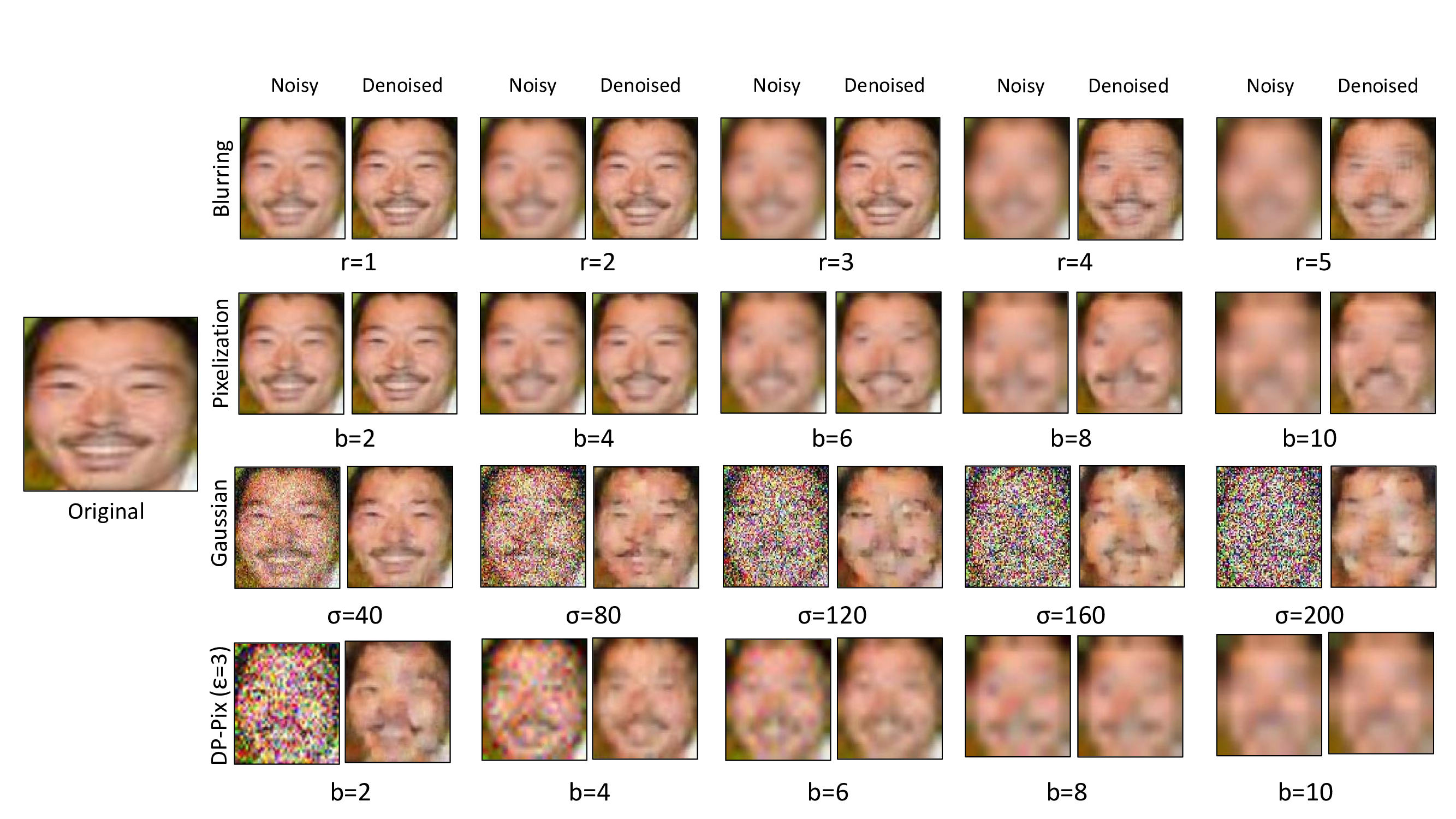}
    \caption{\textbf{Qualitative Evaluation of Privacy Methods:} The figure depicts the obfuscated images (\texttt{Noisy}) by the studied privacy methods at various parameters, as well as the denoised output, with a sample input from CelebA.}
    \label{fig:visual_results_param}
    
\end{figure*}

\textbf{Pixelization:} As shown in Fig.~\ref{fig:reid:pix}, small cell size in pixelization, e.g., $b=2$, leads to high face re-identification rates for both pixelized and denoised images.   Increasing $b$ helps reduce the rate of face re-identification rapidly, e.g., from 53.73\% to 4.82\% for pixelized images by increasing $b$ from 2 to 6.  Denoising slightly increases the privacy risk, but the additional risk diminishes with larger $b$ values and is much lower than observed in blurring.

\textbf{Additive Gaussian:} 
Over the range of $\sigma$ values studied in our experiments, Additive Gaussian inflicts lower privacy risks with a small noise ($\sigma=40$), compared to Blurring and Pixelization.  As shown in Fig.~\ref{fig:reid:gauss}, increasing $\sigma$ leads to a moderate reduction in the privacy risk.  For example, face re-identification rate is 5.11\% at mid noise level ($\sigma=120$), reduced from 8.54\% at low noise level ($\sigma=40$). Denoising the obfuscated images leads to a significant increase in the privacy risk at low $\sigma$ values, e.g., 33.55\% higher in face re-identification rate when $\sigma=40$.

 \textbf{DP-Pix} Fig.~\ref{fig:reid:dp_pix} presents the re-identification results for images obfuscated with DP-Pix as well as those denoised by our proposed approach. We observe low privacy risks across all $b$ values. Furthermore, performing denoising on DP-Pix obfuscated images does not lead to significant higher privacy risks with any $b$ value, as opposed to other image obfuscation methods. While face re-identification rates are consistently low, higher rates occur when $b=4$ and $6$. Recall that higher utility was observed when $b=4$ and $6$ in Fig.~\ref{fig:results}.  It has been reported in~\cite{fan_image_2018} that the quality of obfuscated images may be optimized by tuning $b$ value given the privacy requirement $\epsilon$, by balancing the approximation error by pixelization and the Laplace noise with scale $\frac{255m}{b^2\epsilon}$.

\begin{figure*}[h]
    \centering
    \includegraphics[width=\textwidth]{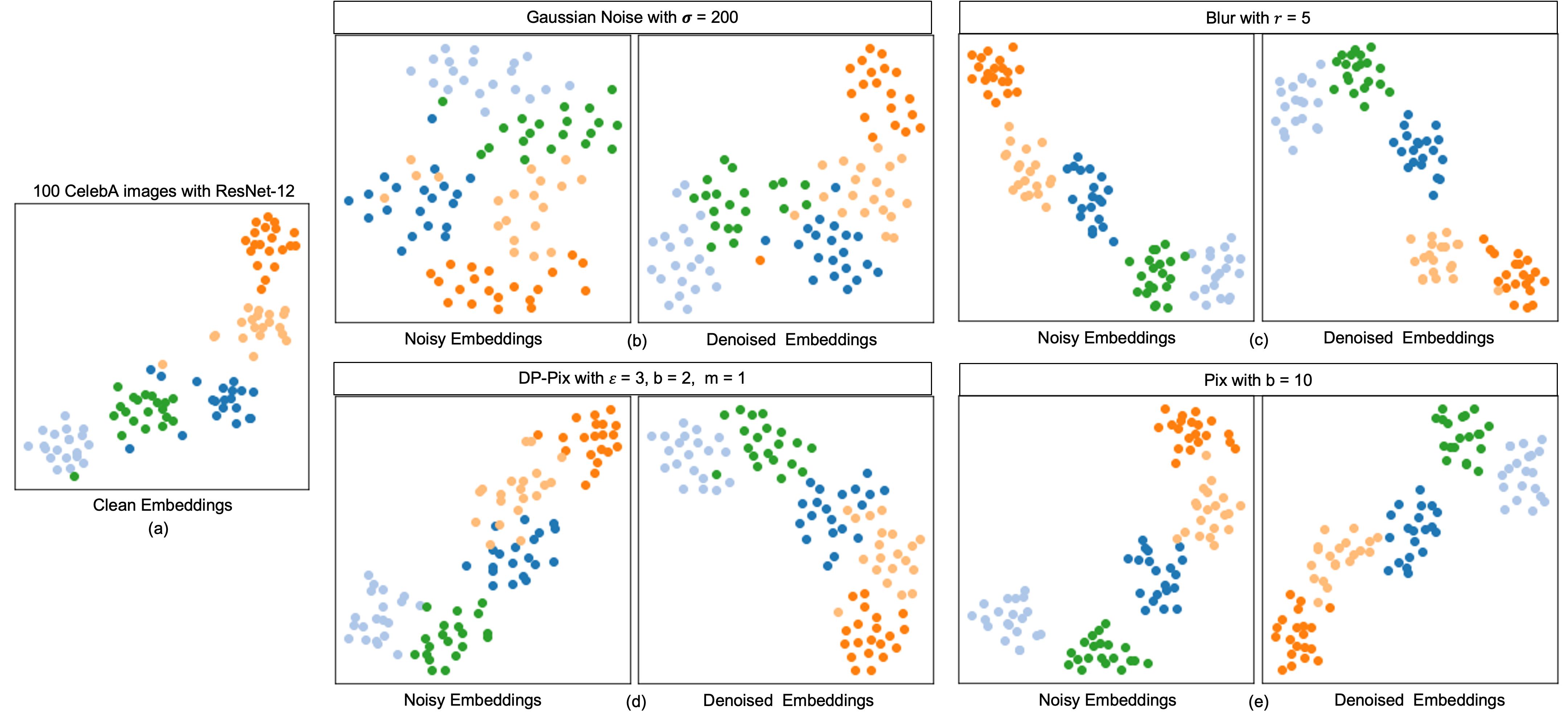}
    \caption{t-SNE Visualization} 
    \label{fig:tSNE}
\end{figure*}

\subsection{Qualitative Evaluation of Privacy Methods}
Fig.~\ref{fig:visual_results_param} provides a qualitative evaluation on the obfuscated and denoised images generated for a range of parameter values. MTCNN \cite{MTCNN} was applied to a sample input image of CelebA, to detect the facial region. Perceptually, the proposed denoiser may improve the image quality upon the obfuscated images to various extents. However, image quality does not always correlate with empirical privacy risks, i.e., face re-identification with public models. In combination with Fig.~\ref{fig:reidentificationresults}, we observe that the proposed denoising leads to various levels of privacy risk increment,  while producing higher quality images.  For example, the results show higher privacy risk increment for Blurring with $r=3$ (23.33\%), moderate increment for Pixelization $b=4$ (10.14\%) and Additive Gaussian $\sigma=80$ (7.96\%), and little increment for DP-Pix $b=2$. Fig.~\ref{fig:visual_results_param} confirms that the denoiser performance may vary depending on the image obfuscation method, and that DP-Pix provides consistent privacy protection even with denoising. Additional qualitative results are presented in Fig. \ref{fig:visual_results} of Appendix.

\subsection{Observation of the Privacy-Preserved Embedding Space}
Fig. \ref{fig:tSNE} shows the evolution of the embeddings in the process of privacy encoding and privacy-preserved representation learning by presenting the t-SNE \cite{van2008visualizing} visualization of the clean, noisy, and denoised embeddings of randomly sampled 100 test images from a total of 5 classes from the CelebA dataset. 
The embeddings are obtained from the ResNet-12 encoder trained under different noise settings for 5-way 5-shot classification. We say the embeddings are clean when the input images have no noise and the encoder is trained for few-shot classification of clean images. The noisy embeddings are obtained by using the encoder trained for few-shot classification of noisy images and without using the denoiser. The denoised embeddings are obtained by the proposed \textit{Denoising Network} (Fig. \ref{fig: cloud}) i.e., the encoder trained in conjunction with denoiser for few-shot classification on noisy images.

We report the results for a case when a few-shot method such as Prototypical Networks can generate good clusters for the clean images (Fig. \ref{fig:tSNE}a), and observe the impact on clustering with noisy images and subsequently when those images are denoised. We notice that when the initial clusters are good, pixelization (Fig. \ref{fig:tSNE}c) and blurring (Fig. \ref{fig:tSNE}e) will have little impact on the quality of the clusters even with the high amount of noise.  Therefore, pixelization and blurring maintain generality (robust to noise) and are also vulnerable to re-identification. Gaussian noise (Fig. \ref{fig:tSNE}b) distorts the initial clusters more significantly, which can lead to lower few-shot classification performance. Applying denoising to Gaussian noise improves the clustering results, however still poses moderate privacy threat as seen in re-identification experiments (Fig. \ref{fig:reid:gauss}). Similarly, with DP-Pix (Fig. \ref{fig:tSNE}d), the original clusters are also distorted upon obfuscation. But, when denoised with proposed Denoising Network,  we can observe better clustering performance.  Because of DP-Pix’s privacy guarantee  and lowest re-identification rates, we can say that the obtained denoised embeddings are privacy-protected i.e., the network finds the privacy-preserved embedding space which maintains generality (robust to noise) and also preserves privacy.

\section{Related Works}
Xie et al. \cite{Xie2020SecureCF} incorporate differential privacy into few-shot learning through adding Gaussian noise into the model training process \cite{abadi_deep_2016} to protect the privacy of training data.
\cite{Xue2021DifferentiallyPP, Huai2020PairwiseLW} have also provided a strong privacy protection guarantee in pairwise learning for training data. On the other hand, \cite{Gelbhart2020DiscreteFL} propose to use hashing to store the embedding of the input images.  Similar to cryptographic approaches~\cite{cabrero2021sok}, the work \cite{Gelbhart2020DiscreteFL} incurs high computational complexity to achieve accuracy.
Differently, our approach addresses the privacy of user data at source (i.e., the images are already privatized before the server sees them) with strong privacy protection. 
To the best of our knowledge, ours is the only approach that addresses privacy in the context of few-shot metric learning for user-supplied training and testing data.

\section{Conclusion \& Future Work}
In this paper, we present a novel framework for training a few-shot private image classification model, which aims to preserve the privacy of user-supplied training and testing data. The framework makes it possible to deploy few-shot models in the cloud without compromising users' data privacy. We discuss and confirm that there exists a privacy-preserved embedding space which has both stronger privacy and generalization performance on few-shot classification. The proposed method provides privacy guarantees while preventing severe degradation of the accuracy as confirmed by results on three different datasets of varying difficulty with several privacy methods. Evaluation with re-identification attacks verifies the low empirical privacy risk of our proposed method, especially with DP-Pix. While our study focuses on well-known image obfuscation methods, future research may explore scenarios where users could apply novel image obfuscation methods locally, i.e., different from those applied to training data.  
Furthermore, our results motivate the future direction of searching for a more effective privacy-preserved space for few-shot learning in other domains such as speech \cite{parnami2020few}. Examination of other evaluation metrics for privacy-preserved embedding space will promote the relevant future study. We release the code for our experiments at \url{https://github.com/ArchitParnami/Few-Shot-Privacy}.

\section*{Acknowledgement}
This work has been supported in part by the National Science Foundation CNS-1949217, CNS-2003198 and UNC Charlotte. The opinions, findings, and conclusions or recommendations expressed in this material are those of the authors and do not necessarily reflect the views of the sponsors. 

\bibliography{references}

\pagebreak
\onecolumn
\appendix

\begin{figure}[ht]
    \centering
    \includegraphics[width=\textwidth]{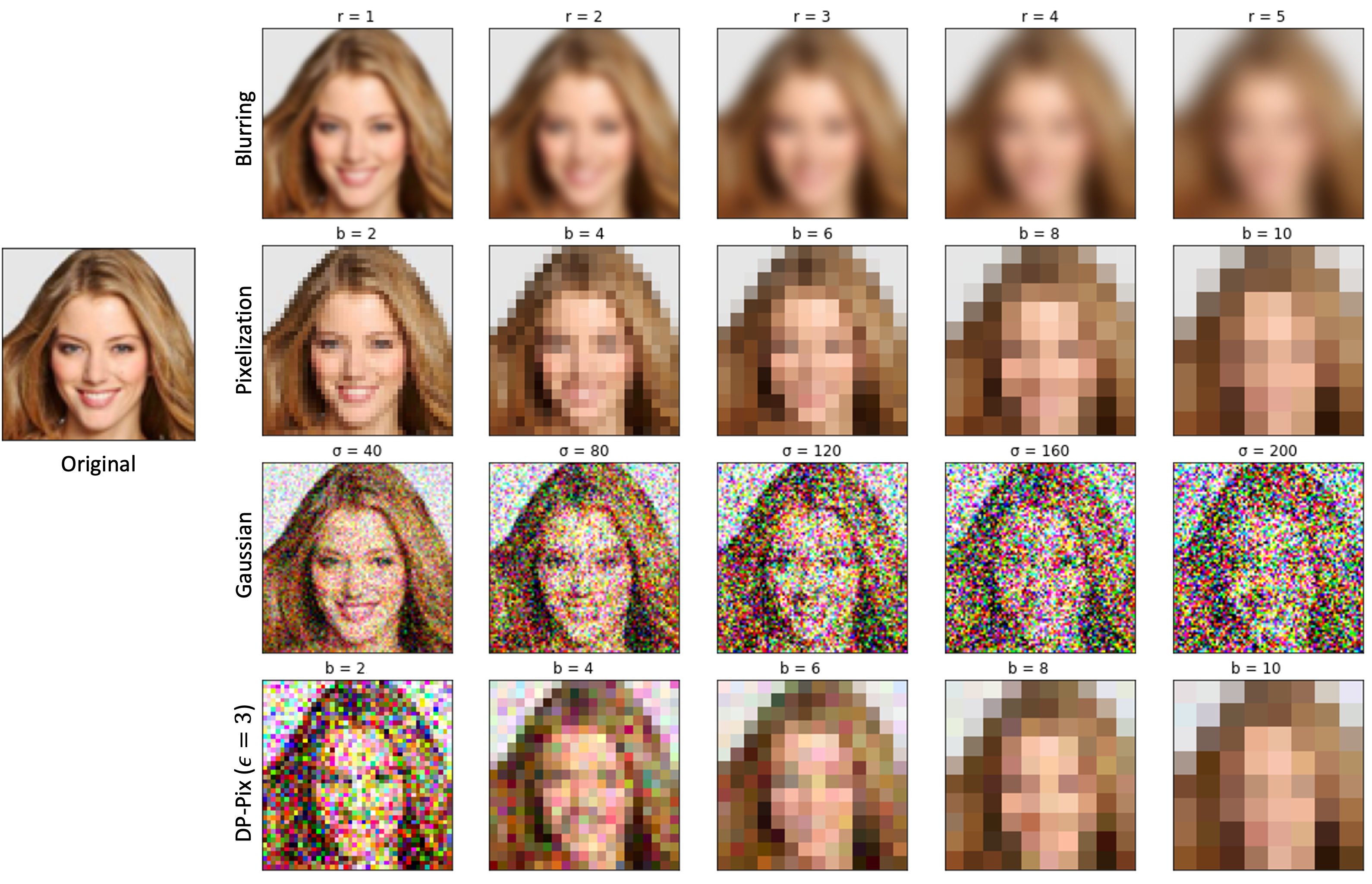}
    \caption{Privacy Methods (Original image from CelebA dataset)}
    \label{fig:privacy_methods}
\end{figure}

\pagebreak

\begin{figure}
    \centering
    \includegraphics[width=\textwidth]{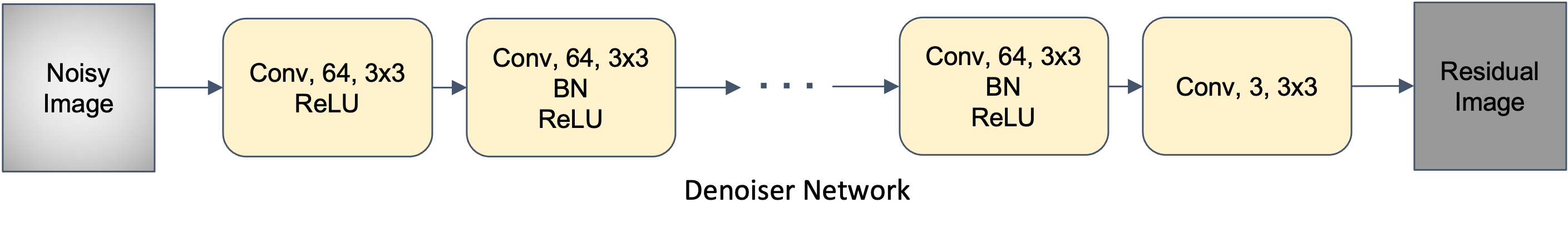}
    \caption{DnCNN denoiser with 8 layers}
    \label{fig:denoiser}
\end{figure}

\vspace{2cm}

\begin{figure}
    \centering
    \includegraphics[width=\textwidth]{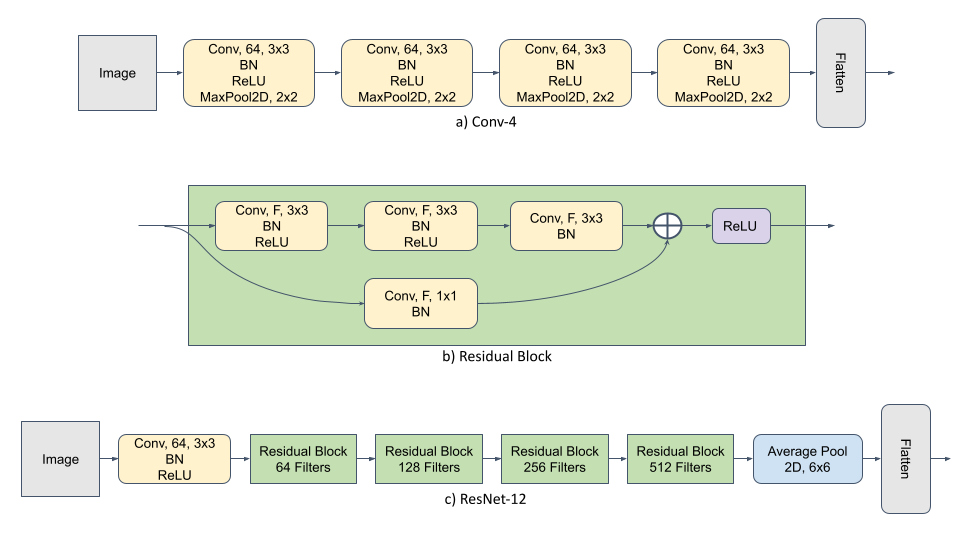}
    \caption{Encoders: a) Conv-4, b) Residual Block, c) ResNet-12}
    \label{fig:encoders}
\end{figure}

\begin{figure}
    \centering
    \includegraphics[width=\textwidth]{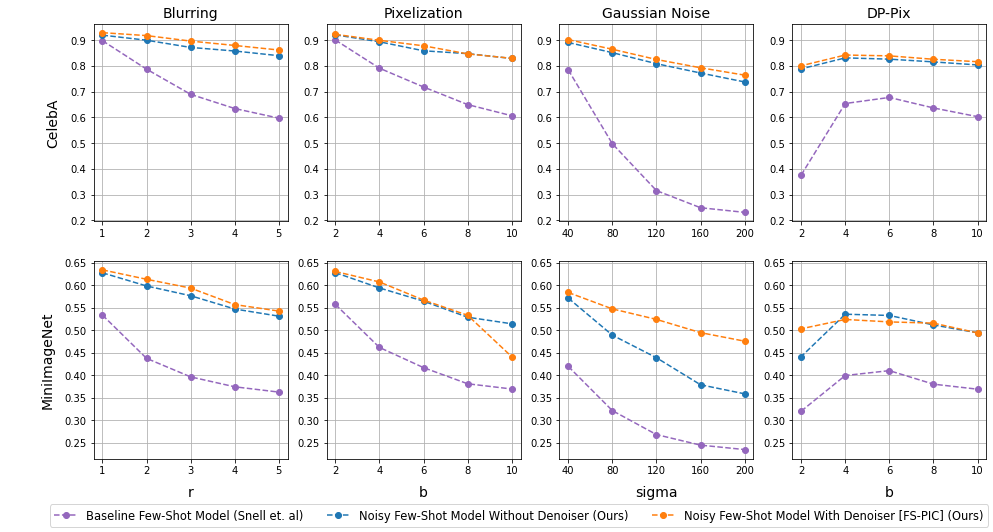}
     \caption{Test accuracy (y-axis) of 5-way 5-shot private image classification tasks sampled from CelebA (top) and MiniImageNet (bottom) datasets, presented with different privacy settings (x-axis) when using \textbf{ResNet-12} as encoder.}
    \label{fig:results_resnet}
\end{figure}

\begin{figure}
    \centering
    \includegraphics[width=\textwidth]{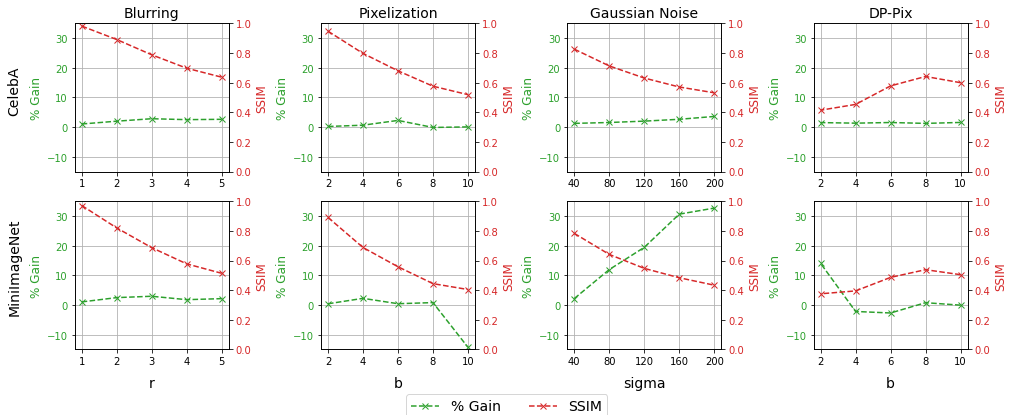}
    \caption{\%Gain vs SSIM for \textbf{ResNet-12}}
    \label{fig:gain_resnet}
\end{figure}

\begin{figure}
    \centering
    \includegraphics[width=\textwidth]{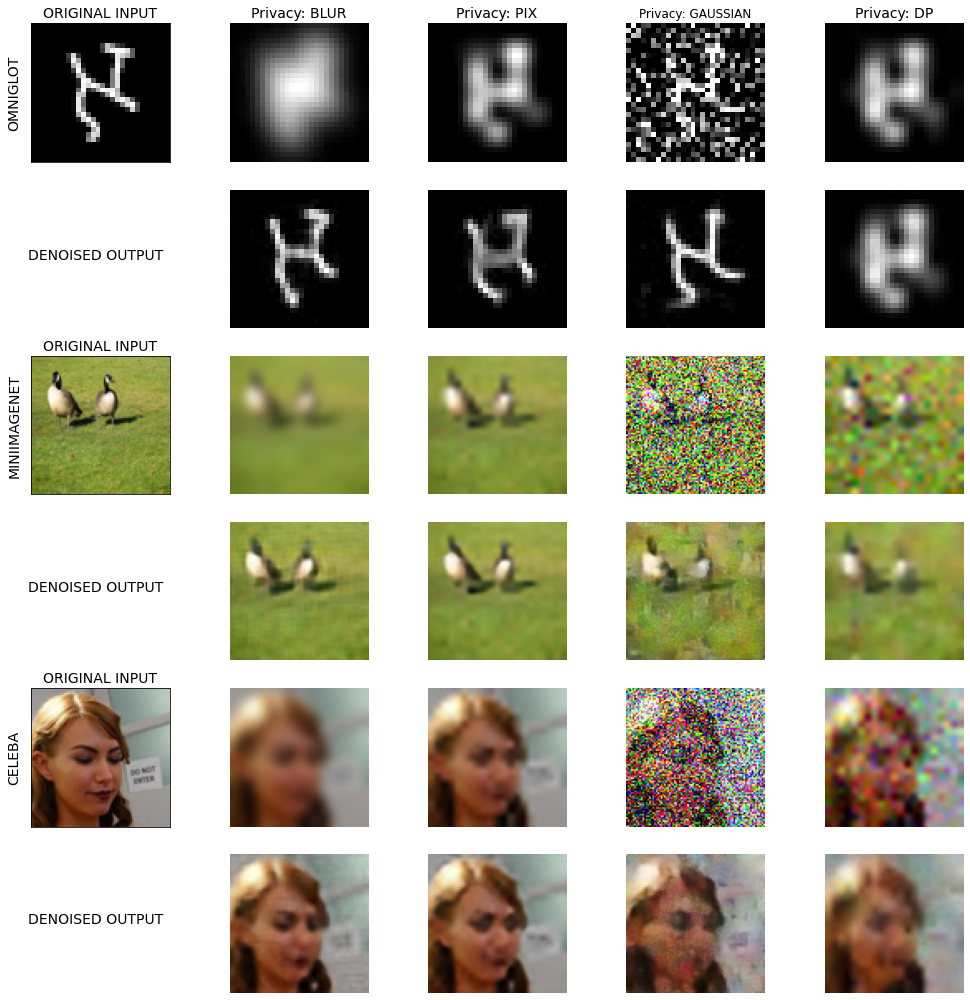}
    \caption{The figure shows the results of applying different types of privacy methods i.e., Blurring, Pixelization, Additive Gaussian noise, and Differentially Private noise to a sample from each of Omniglot, MiniImageNet and CelebA datasets. It also shows the denoised output obtained from the DnCNN denoiser. We observe that images protected with DP-Pix are hard to denoise when compared with Blurring, Pixelization, and Gaussian noise.}
    \label{fig:visual_results}
\end{figure}

\end{document}